\documentclass[10pt,twocolumn]{article}
\usepackage{abstract}
\usepackage[utf8]{inputenc}
\usepackage[T1]{fontenc}
\usepackage{lmodern}
\usepackage{microtype}
\usepackage{graphicx}
\usepackage{caption}
\usepackage{subcaption}
\usepackage{amsmath,amssymb}
\usepackage{booktabs}
\usepackage{authblk}
\usepackage{geometry}
\usepackage[style=nature, sorting=none, maxbibnames=5]{biblatex}
\usepackage{xurl}

\usepackage{color, colortbl}
\definecolor{Gray}{gray}{0.9}

\usepackage{sectsty}
\allsectionsfont{\normalfont\bfseries}

\title{\Large\bfseries The impact of tactile sensor configurations on grasp learning efficiency - a comparative evaluation in simulation}
\author[1]{Eszter Birtalan\thanks{Corresponding author: birtalan.eszter@itk.ppke.hu}}
\author[1]{Miklós Koller}
\affil[1]{Faculty of Information Technology and Bionics, Pázmány Péter Catholic University, Budapest, Hungary}

\renewenvironment{abstract}{
  \small
  \begin{center}\bfseries Abstract\end{center}
  \begin{quote}\noindent
}{
  \end{quote}
}

\begin{document}
\twocolumn[
  \date{}
  
  \maketitle
  
  \begin{@twocolumnfalse}
  \begin{abstract}
  \noindent
Tactile sensors are breaking into the field of robotics to provide direct information related to contact surfaces, including contact events, slip events and even texture identification. These events are especially important for robotic hand designs, including prosthetics, as they can greatly improve grasp stability. Most presently published robotic hand designs, however, implement them in vastly different densities and layouts on the hand surface, often reserving the majority of the available space. We used simulations to evaluate 6 different tactile sensor configurations with different densities and layouts, based on their impact on reinforcement learning. Our two-setup system allows for robust results that are not dependent on the use of a given physics simulator, robotic hand model or machine learning algorithm. Our results show setup-specific, as well as generalized effects across the 6 sensorized simulations, and we identify one configuration as consistently yielding the best performance across both setups. These results could help future research aimed at robotic hand designs, including prostheses.
  \end{abstract}
  \vspace{0.6cm}
  \end{@twocolumnfalse}
]
\saythanks
\begin{figure*}[h]
  \includegraphics[width=1\textwidth]{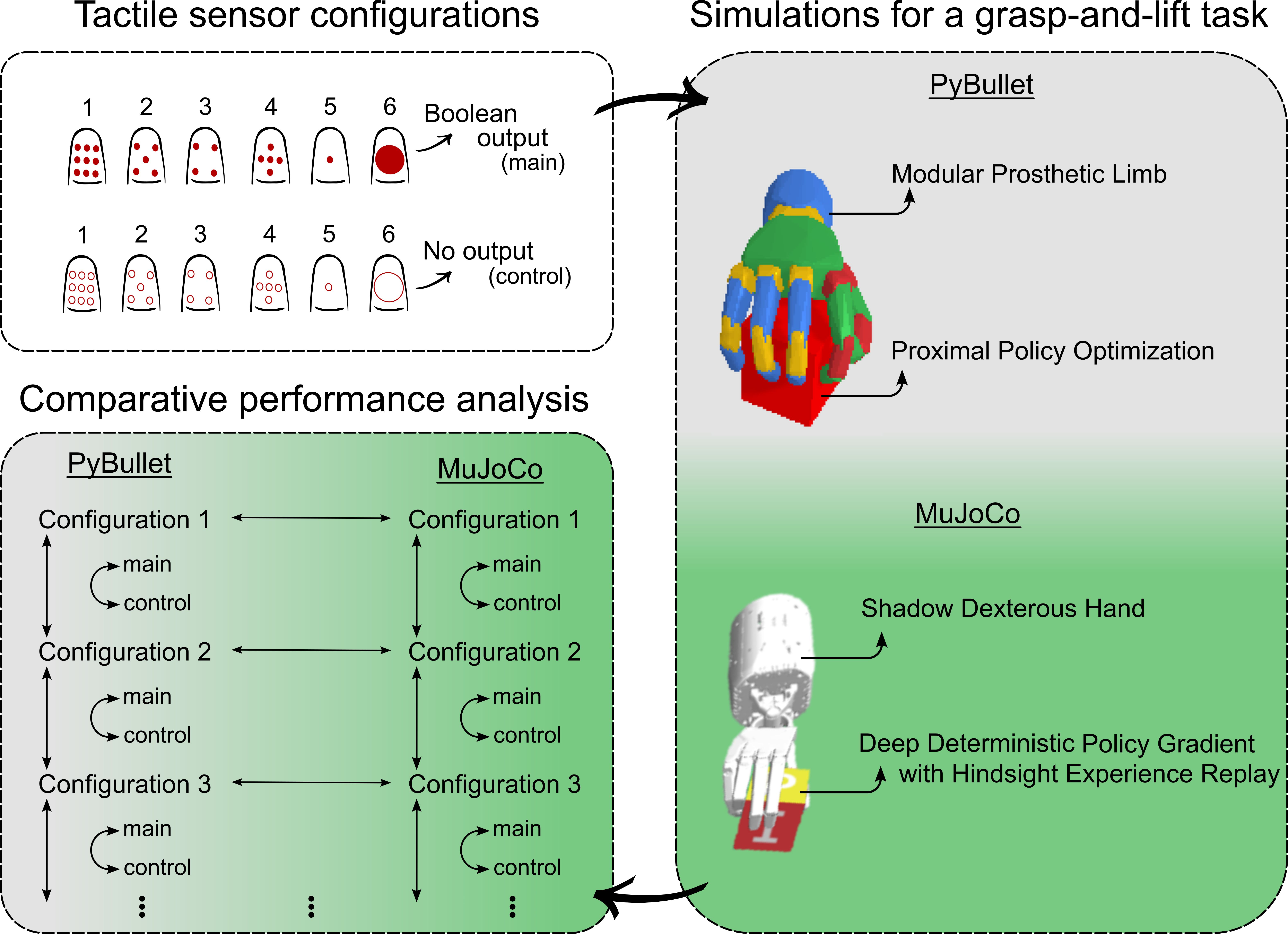}
  \caption{\textbf{Schematic overview of the methodology used to evaluate tactile sensor configurations.} The same configurations have been implemented on two 3D hand models using different physics simulators. The effects of the different sensor layouts were evaluated using two reinforcement learning algorithms to ensure the robustness of the results and compared against configurations, setups and control groups.}
  \label{fig:fig1}
\end{figure*}

\section*{Introduction}
Dexterous manipulation and grasping have been captivating fields of research in robotics, as they are useful both in autonomous robots made for complex task solving and also in the field of robotic prosthetics. Object manipulation, especially in the dark, is still a task that humans perform significantly better than robots do. Early approaches to solve this problem aimed to describe the exact model of contact, position, and motion parameters, as well as the control framework and motion planning, which worked reliably and well for low complexity environments and tasks \cite{sundaralingam_relaxed-rigidity_2017}. 

With the increase in complexity, other solutions relying on machine learning, such as deep learning and deep reinforcement learning, have become more widespread \cite{xu_towards_2022, valarezo_anazco_natural_2021, andrychowicz_learning_2020}. Unlike with previous approaches, here the exact state of the robot is not known, but may be inferred by the use of sensors. Traditionally, visual sensors and proprioceptive information have been used for state estimation \cite{shi_computer_2020}. In more recent works, however, tactile sensing for object interaction has been shown to be of great value \cite{melnik_using_2021} in addition to other modalities. For example, visual feedback may be used for state estimation while tactile sensors are used for slip detection and correction \cite{yu_realtime_2018, motamedi_haptic_2016}, or the combination of proprioceptive and tactile data may be used to boost learning and performance for object manipulation tasks, foregoing visual information altogether \cite{melnik_using_2021}. A notable advantage of tactile sensors is their variability and adaptability for various environments. Sensor mechanisms may range from simple piezo-electric \cite{navarai_capacitive-piezoelectric_2018, navaraj_prosthetic_2019, cotton_novel_2007}, piezo-resistive \cite{castellanos-ramos_tactile_2009}, optical sensors \cite{james_slip_2018, yuan_gelsight_2017}, to complex sensor designs with intricate 3D structures \cite{kampmann_integration_2014} measuring light scattering to derive information about contacts. Modern designs also allow for high-density solutions involving sensor arrays with conductive ink \cite{castellanos-ramos_tactile_2009}, or vision-based contact sensing, transforming up to 70\% of the palmar side of the hand to a sensing surface \cite{zhao_embedding_2025}.

With limited space on a robotic hand, serious considerations need to be made when choosing between the available sensor types. For autonomous robots, high coverage and high density solutions seem to be preferred \cite{zhao_embedding_2025, schmitz_methods_2011, fukui_high-speed_2011, mouri_anthropomorphic_2002}; however, in the case of robotic prosthetics, low-resolution, array-like, or even single-point tactile sensing could be more advantageous \cite{cipriani_smarthand_2011, osborn_tactile_2014,navarai_capacitive-piezoelectric_2018, navaraj_prosthetic_2019, abbass_embedded_2021, zhang_design_2018, Maggiali08}. It is important to note that in nearly all of these examples, both the coverage and the number or positioning of contact sensing elements vary greatly, with no clear consensus. \\
Navaraj et al. \cite{navarai_capacitive-piezoelectric_2018, navaraj_prosthetic_2019} developed a novel tactile sensor capable of both static and dynamic sensing, mimicking the fast- and slow-adapting mechanoreceptors of the human hand. One of these sensors covered most of the area of a single phalanx and has not been implemented on the palm.\\
Cipriani et al. \cite{cipriani_smarthand_2011} included only 4 tactile sensors in total in their hand design, focusing them on the area of the hand that interacts most with objects: the thumb and index fingers. \\
The flexible matrix electrode developed by Abbass et al. \cite{abbass_embedded_2021} may contain 4, up to 16 sensors, which were screen-printed onto a flexible substrate. Their proposed sensor distribution on a prosthetic hand included 7 sensors on the index fingertip, and 4-4 sensors on the middle and proximal phalanges in order to participate in sensory feedback for the user.\\
Zhang et al. \cite{zhang_design_2018} developed their own prosthetic hand, equipped with 13 contact sensing units aimed at selecting appropriate grasping force and performing slip detection for a more stable grasp. The tactile sensing occurred on fingertips only, which contained all 13 sensing units. \\
Osborn et al. \cite{osborn_tactile_2014} focused on developing a new type of tactile sensing that could create a low-cost, adaptable sensor array designed specifically for use on any type of prosthetic limb. The sensing cuffs could be made to any dimensions, achieving a variety of spatial resolutions based on need. The stretchable cuffs could be attached to the phalanges and to the palm as well, resulting in high coverage across the hand.

Unlike in autonomous robots or more general robotic-arm designs and simulations, tactile sensors in prosthetic hands focus more on sensory feedback or slip detection, however, it would be feasible to involve them in grasp generation as well. To facilitate this, we investigated the effect of tactile sensor density and layout on reinforcement learning using hand models that may be used as prostheses (Modular Prosthetic Limb \cite{johannes_chapter_2020} and Shadow Dexterous Hand \cite{shadowrobot2013}). These investigations could allow future research to direct fewer resources into sensor layout design, and avoid potential dead-ends. In this research we focused specifically on grasping motions, as these are commonly offered by robotic prosthetic limbs, and evaluated configurations of up to 9 sensors per phalanx, to keep possible implementation costs low. Aside from the 9-sensor configuration, all our other layouts permit space for other sensors to be possibly implemented, much like in the works of Cipriani et al.\cite{cipriani_smarthand_2011}.\\
A possible hurdle that may affect such projects with machine learning is that simulation results may be applicable only to the exact setup used. Either due to the physical parameters of the models, the inner workings of the physics simulator, the machine learning algorithm, or other, unknown factors. To make our results more robust, we created two simulated setups, which differ in every aspect except for the task, sensor configurations, and sensor output (Fig. \ref{fig:fig1}). The first setup, created using the PyBullet \cite{pybulletcoumans2016} physics simulator, included the Modular Prosthetic Limb (MPL) and a Proximal Policy Optimization (PPO) algorithm \cite{schulman_proximal_2017}. The second setup, made in the MuJoCo \cite{todorov_mujoco_2012} physics simulator, included the Shadow Dexterous Hand (Shadow Hand), and a combination of Deep Deterministic Policy Gradient \cite{lillicrap_continuous_2019} and Hindsight Experience Replay \cite{andrychowicz_hindsight_2018} (DDPG+HER), based on the works of Melnik et al. \cite{melnik_using_2021}. The task of grasping and lifting a small cube remained the same across the setups, as did the layouts of the tactile sensors and their Boolean-type output. We thought it imperative that the task should include a hand model facing palm down, as in such a case, any grasping technique that wasn't forming a stable grip would result in a failed task due to the object dropping. To mitigate the variability in learning performance as a result of this type of experimental setup, we used bootstrapping methods to analyze our data. Our results show that it is possible to positively benefit from tactile sensors using a lower-resolution layout, thus allowing space to include other sensor modalities in the prosthetic device, and that sensor layout affects learning results significantly, even if the number of sensors used remains the same.

\begin{figure}[h]
  \centering
  \includegraphics[width=0.75\columnwidth]{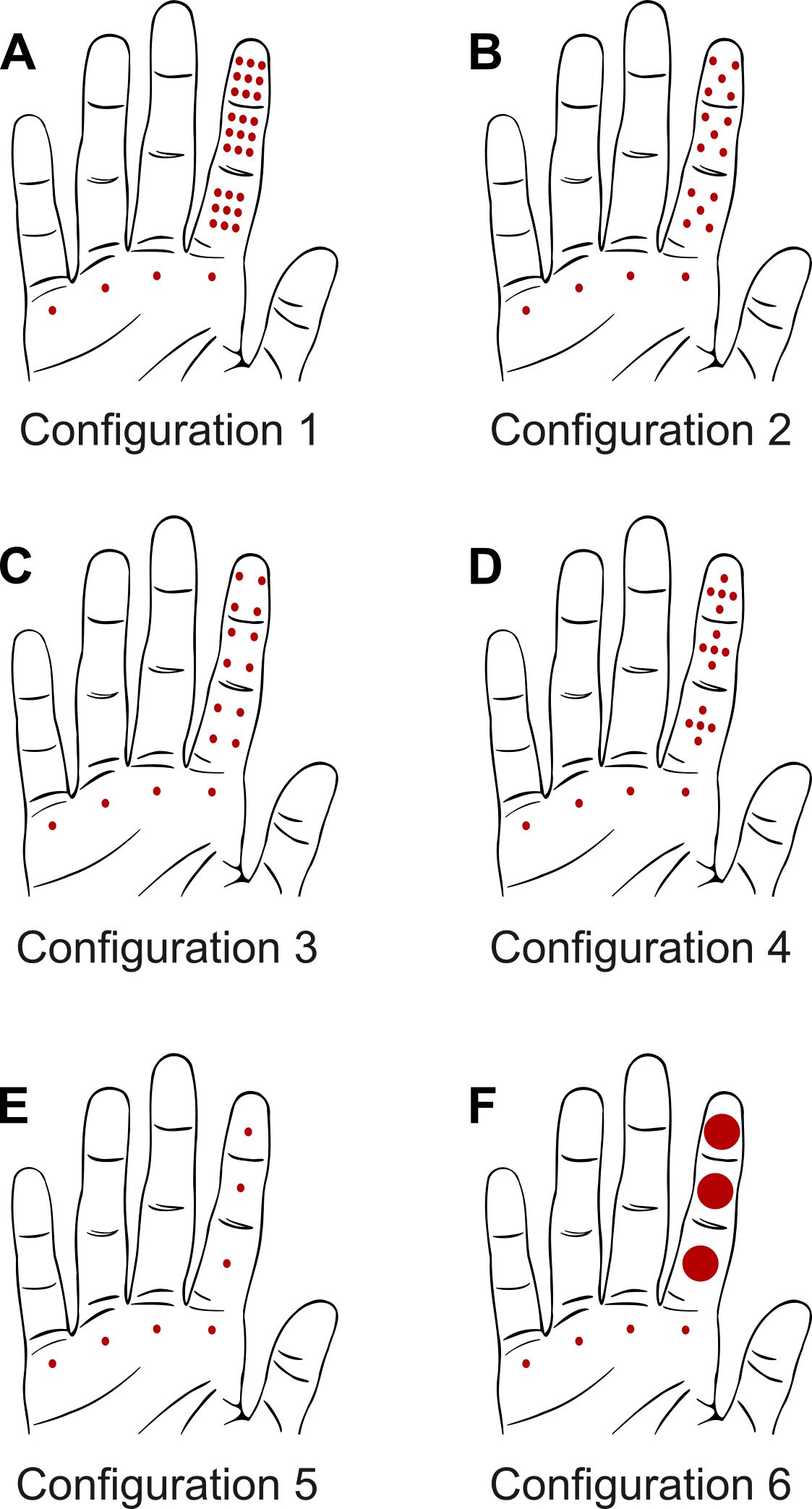}
  \caption{\textbf{Visual representation of the sensor configurations implemented in both setups.} A-F: Sensor configurations corresponding to Configurations 1-6, respectively. The number and layout of the sensors shown on the index finger are repeated across the other digits.}
  \label{fig:fig2}
\end{figure}

\section{Results}
\begin{figure*}[!h]
  \centering
  \includegraphics[width=0.85\textwidth]{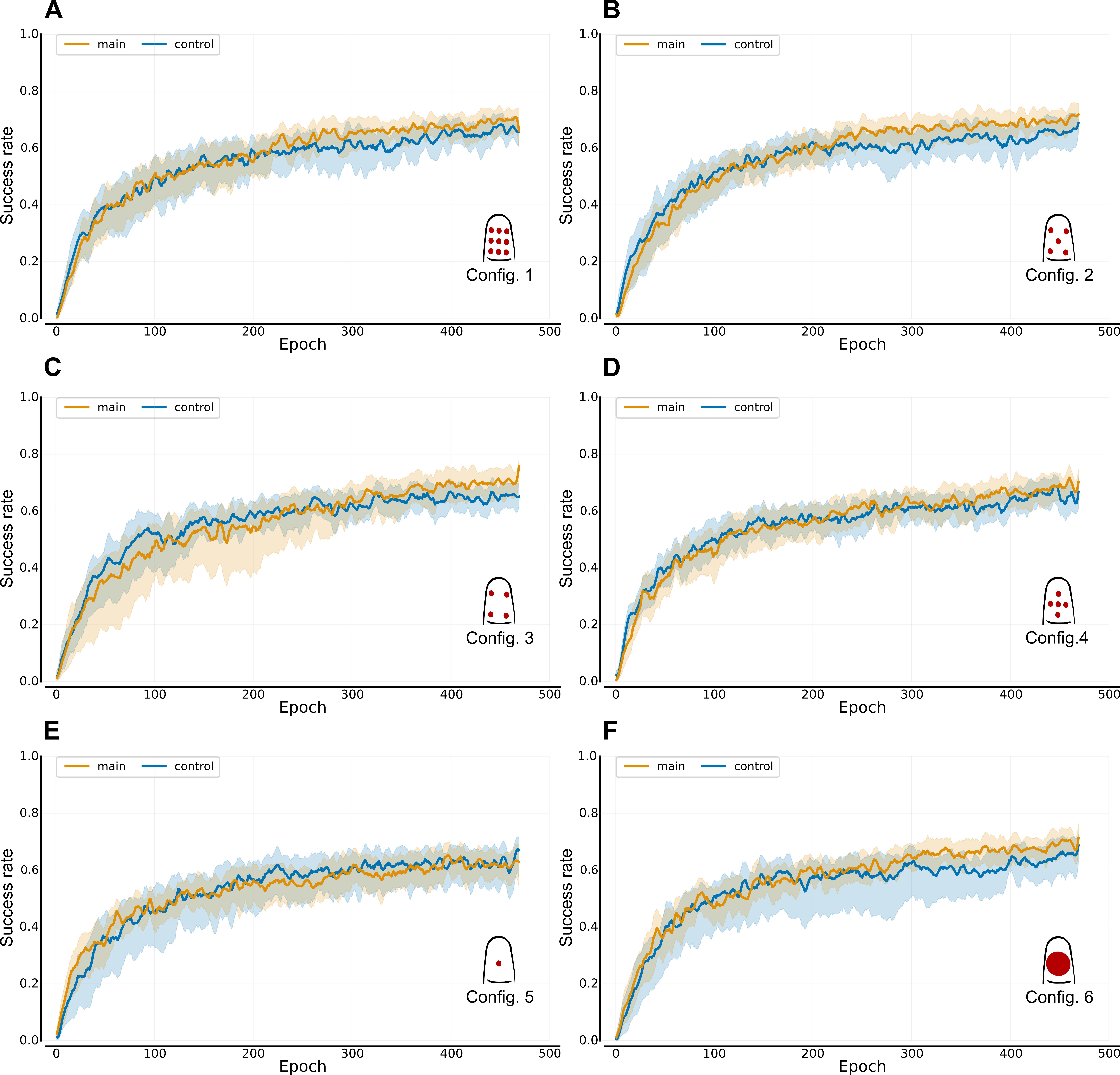}
  \caption{\textbf{Sample-efficiency curves showing success rates as a function of epochs in the PyBullet setup.} A-F: Sample-efficiency curves corresponding to Configurations 1-6 (shown on diagrams in the right hand corner of each panel), respectively. Thick lines show the IQM scores (bootstrap replication 50000), with the confidence intervals represented as shaded areas. Yellow is used to show results from the sensorized group, while blue is used for the controls. Training occurred for 500000 timesteps, roughly generating 500 epochs. Both main and control groups represent the result of 10 seeds.} 
  \label{fig:fig3}
\end{figure*}

We evaluated the effects of sensor placement on reinforcement learning by establishing two simulated setups, where each agent learns to grasp and lift a cube using different hand models and learning algorithms, while keeping the sensor configurations the same. During reinforcement learning, each training episode begins with the object (a cube) placed directly under the palm of the hand model, which has its fingers in a slightly curved position. The agent needs to learn to establish a stable grasp around the object, that prevents a drop when the hand is lifted. If the grasp proves to be unsecure, the object will fall to the ground, and the task cannot be completed. This way the agent has a much lower chance to learn semi-secure holds, at the cost of larger deviations in task success. We train with 6 different sensor configurations (Figure \ref{fig:fig2}) to observe the effects of different sensor densities, sensor locations (Figure \ref{fig:fig2} B and D), sensing in the middle of the phalanx (Figure \ref{fig:fig2} C and E), and finally a single sensor design (Figure \ref{fig:fig2} F). Each configuration is complemented with 4 sensors on the palm, placed at the base of each finger, excluding the thumb. Thus, Configurations 1-6 have a total of 139, 79, 64, 79, 19, and 19 tactile sensors, respectively. In the presence of active sensor measurements (called main run type), the sensors return Boolean values based on activation, while in absence of active measurements (no real tactile data, this is called control run type) these values are ’no contact detected’ for each query. With this control scheme, we ensure that the input layer of the neural network always receives a matrix of the same size within configurations, regardless of run type (main or control).\\
Learning results are evaluated through success-rate measurement, which calculates the ratio of successful episodes to all episodes for each epoch (PPO) or each test cycle (DDPG+HER), based on the algorithm. We use the RLiable Python library to compute stratified bootstrap with 95\% confidence intervals for our aggregated performance metrics (Median, Interquartile Mean - IQM, and Mean for converged success-rates) across seeds, following the methodology of Agarwal et al. \cite{agarwal_deep_2022}. Converged performance is calculated according to Melnik et al. \cite{melnik_using_2021}, by transforming the learning curves to histograms projected onto the Y axis and taking the performance value paired with the largest bin of the histogram. We also include sample-efficiency curves generated by RLiable to visualize the learning curves for each configuration by presenting bootstrapped IQM scores as a function of epochs.
Due to the long training time, and several configurations used, we were unable to produce large sample sizes. Mitigation of issues arising from this sample size was our reasoning behind choosing the stratified bootstrap method. This way, we can resample our data to generate thousands of resampled versions from a single case of Configuration and run-type combination and calculate the main metrics (median, IQM, mean) for each. Then we calculate the confidence interval (CI) by keeping only the central 95\% of the bootstrap distribution. This reduces the influence of rare resampling artifacts while providing a robust estimate of the variability of the performance metrics under repeated experiments. We use these intervals to qualify how well our main metrics represent the expected performance, with small CIs representing stable, and wide CIs representing volatile results, where repeated experiments may yield a wide range of performances. We also use the overlap between these CIs to measure how well the performance of two groups separate from each other to allow qualitative and indicative comparisons. This is especially crucial as data from reinforcement learning is often non-symmetric and non-parametric, making classical statistical comparisons poorly suited. 

\subsection{PyBullet results}
Results of learning over time in the PyBullet setup show that initial learning is negatively affected by the lack of the central sensor when using Configuration 3  (Figure \ref{fig:fig3} C) with tactile information. Other than this, initial learning does not appear to differ between sensorized (main group) and control versions (control group). After the initial phase, however, a distinct separation can be observed between runs with and without tactile information. In the cases of Configurations 1, 2, and 6, runs with tactile information perform consistently better than control versions (Figure \ref{fig:fig3} A, B, F), whereas with Configuration 3, this effect shows up delayed, affecting only the very end of training (Figure \ref{fig:fig3} C). Runs with Configurations 4 and 5 appear to perform similarly regardless of the addition of tactile information or not (Figure \ref{fig:fig3} D, E). Based on the results of the learning curves, we conclude that adding tactile information resulted in better overall performance in 3 of the 6 configurations (Configurations 1, 2, and 6), and that the missing central tactile sensor in Configuration 3 caused a noticeable negative effect on learning.

\begin{figure}[h]
  \centering
  \includegraphics[width=0.50\textwidth]{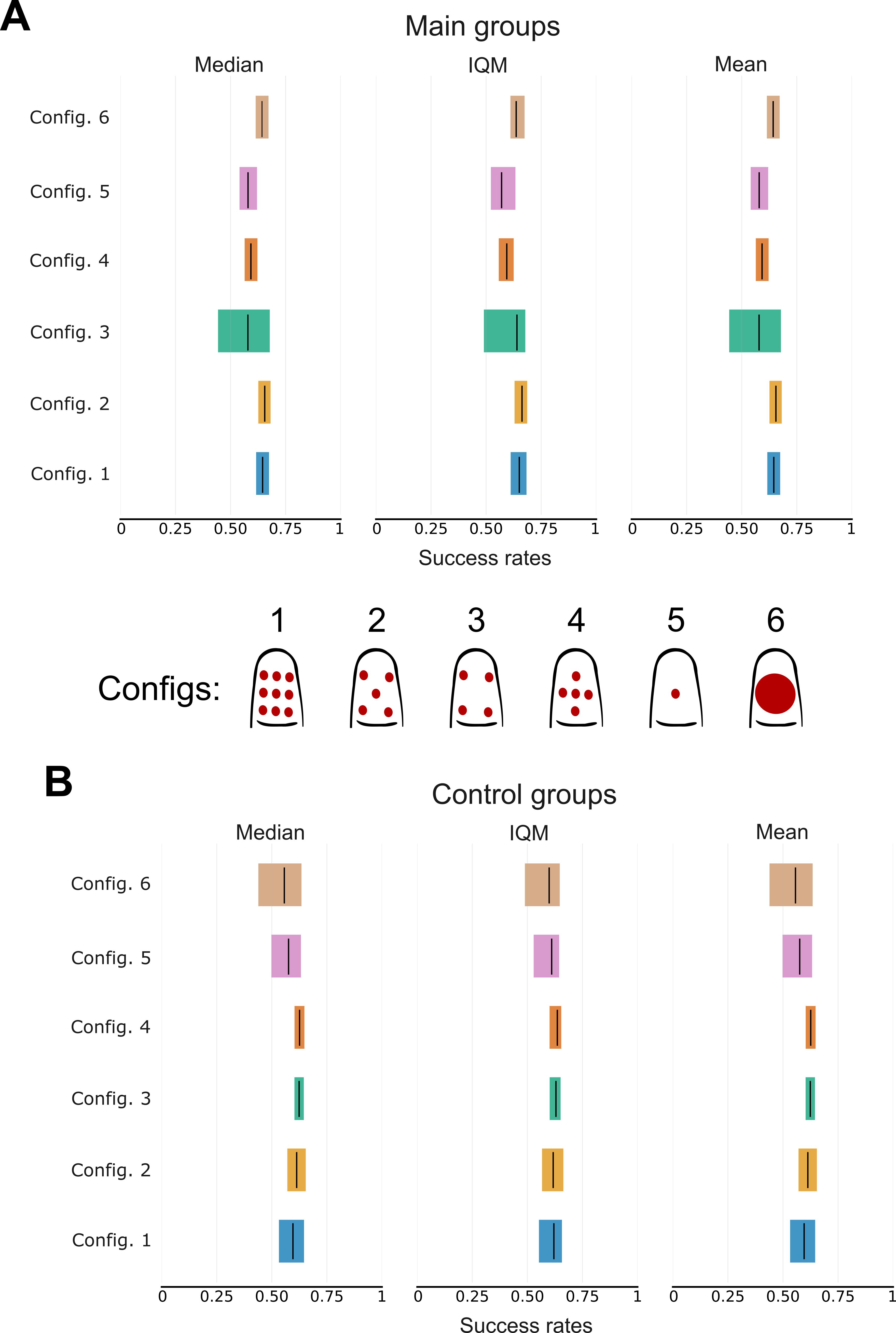}
  \caption{\textbf{Visual representation of the converged success rates in the PyBullet setup.} A-B: The Median, IQM and Mean of the converged success rates (bootstrap replication 1000) for each Configuration in the sensorized and control groups, respectively. Confidence intervals are represented as shaded areas. Each configuration was run with 10 seeds; the exact values represented here can be found in Table \ref{tab:table1}.}
  \label{fig:fig4}
\end{figure}

The bootstrapped converged metrics for both sensorized and control versions are shown in Figure \ref{fig:fig4}, and their values are collected in Table \ref{tab:table1}. The RLiable library calculates the Interquartile Mean (IQM) by discarding the lowest and highest 25\% of the data to provide a clearer picture of performance, which is not skewed by large outliers. In this next section, we will take a look at how the converged performances compare between sensorized and control, and also between the various sensorized versions, using IQM values and their confidence intervals (CIs).

Using Configuration 1, the main group achieved a success rate of 65\%, representing a 3\% improvement over the control, with only 38\% overlap between the corresponding confidence intervals (CIs). CI overlap is always calculated relative to the control, meaning what percentage of the control group's CI overlaps with the main group's CI. With a similar CI overlap of 39\%, the main group of Configuration 2 converged to a 66\% success rate compared to 61\% for the control. The CI for the control group of Configuration 3 fell entirely within the CI of the main group, which had a 1\% higher success rate. The main group of Configuration 4 achieved a success rate of 59\%, compared to 64\% for the control, with a 52\% overlap in CIs. Configuration 5 had similar results, with a 93\% overlap between CIs, achieving 57\% and 61\% success rates for the main and control groups, respectively. Finally, using Configuration 6, the main group performed with a 64\% success rate, 4\% higher than the control group, with a 24\% overlap in CIs compared to the control. \\
Comparing the results from the main groups shows that despite the different number of sensors used in Configuration 1 (139) and Configuration 2 (79), the agent achieved similar success rates (65 and 66\%, respectively), with similarly sized CIs (6-7\%). The performance of Configuration 3, with the missing middle sensor, was not far off with a 64\% success rate; however, its CI was the highest among all the groups, 18\%. Configuration 4 had the same amount of tactile information as Configuration 2; however, the different sensor layout resulted in a lower, 59\% success rate. The size of its CI was 7\%, only 1\% off from Configuration 2. 
Configurations 5 and 6 had a single central sensor on each phalanx, with different activation area sizes. Configuration 5, with the smaller activation area, performed with the lowest success rate among the main groups, and had the second highest CI, 57\% and 11\%, respectively. Configuration 6, with the larger activation area, on the other hand, performed close to Configurations 1 and 2 with a 64\% success rate and an 8\% CI.

\begin{figure*}[!h]
  \centering
  \includegraphics[width=0.80\textwidth]{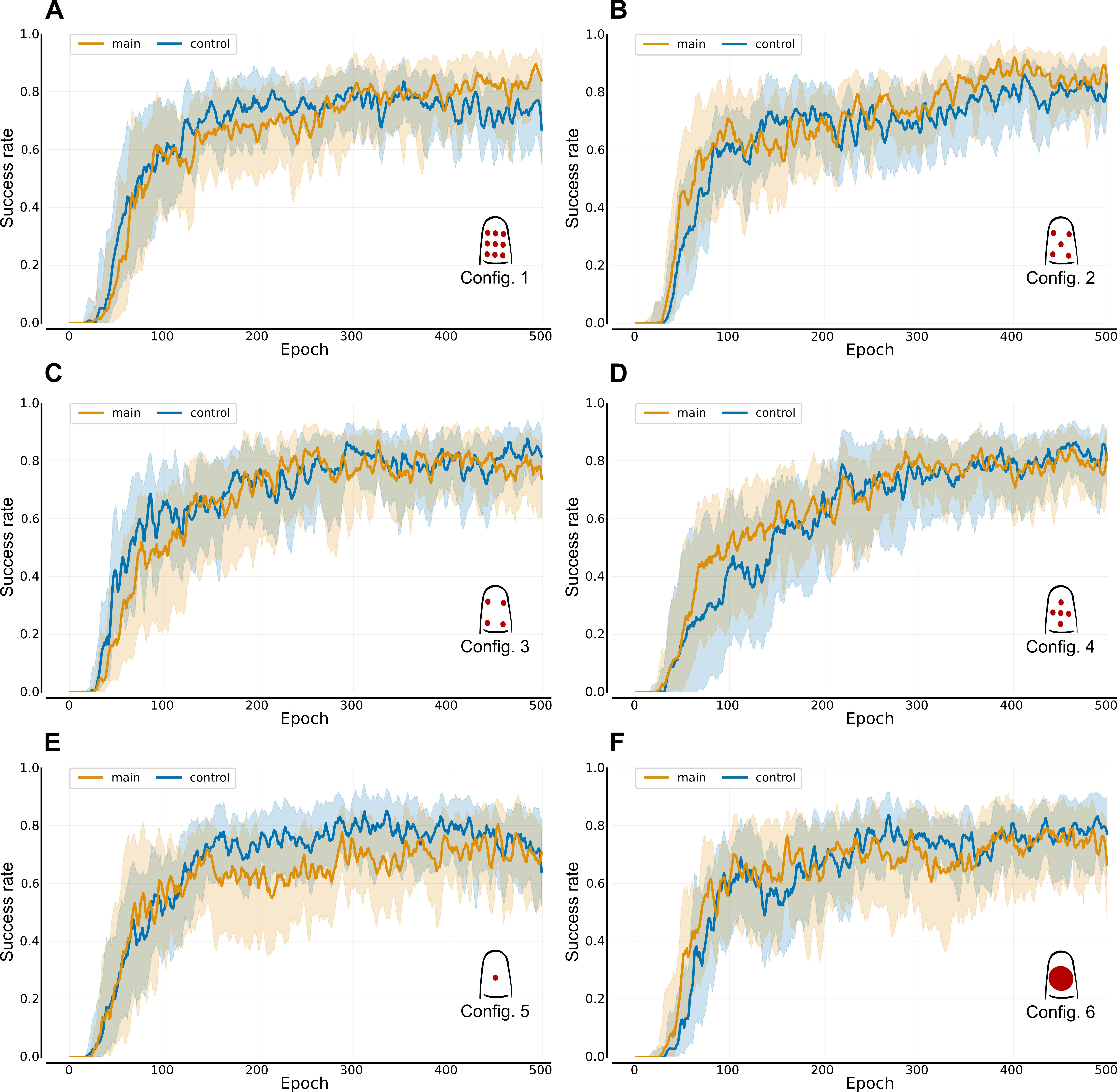}
  \caption{\textbf{Sample-efficiency curves showing success rates as a function of epochs in the MuJoCo setup.}  A-F: Sample-efficiency curves corresponding to Configurations 1-6 (shown on diagrams in the right hand corner of each panel), respectively. Thick lines show the IQM scores (bootstrap replication 50000), with the confidence intervals represented as shaded areas. Yellow is used to show results from the sensorized group, while blue is used for the controls. Training occurred for 5000000 timesteps, generating 500 epochs. Both main and control groups represent the result of 10 seeds.} 
  \label{fig:fig5}
\end{figure*}

\subsection{MuJoCo results}
Results of the MuJoCo environment, paired with a DDPG+HER algorithm, generated larger CIs (16-30\%) than the PyBullet environment with PPO. Therefore, learning tendencies based on the efficiency plots (Figure \ref{fig:fig5}) can be inferred in a more limited manner. Runs with tactile information still outperform control versions at the end of training when using Configurations 1 and 2 (Figure \ref{fig:fig5} A,B), whereas Configuration 5 shows a great reduction in the main group's performance(Figure \ref{fig:fig5} E). Configurations 2, 3, and 6, on the other hand, do not show clear tendencies for group separation (Figure \ref{fig:fig5} B, C, F).

\begin{figure}[h]
  \centering
  \includegraphics[width=0.5\textwidth]{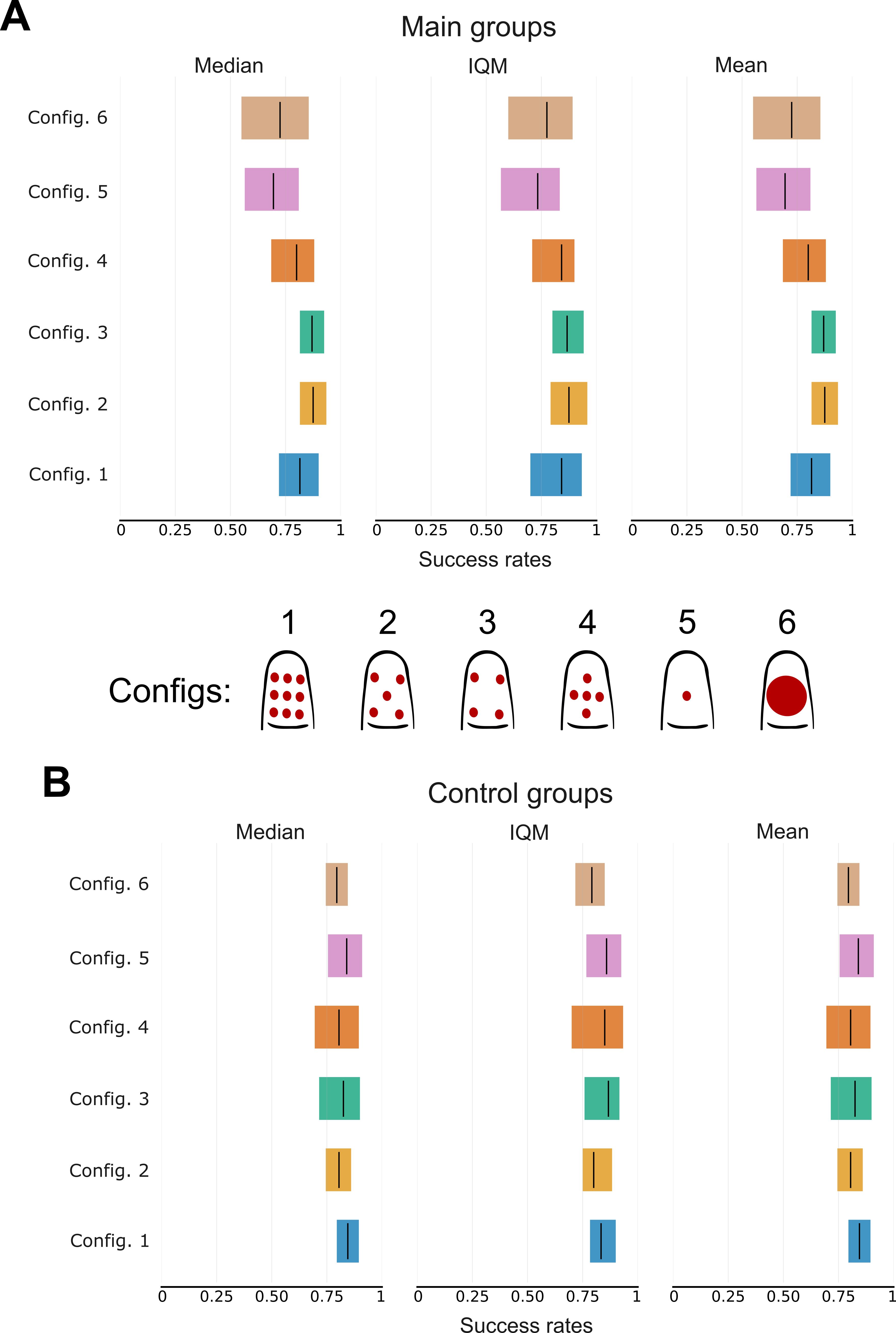}
  \caption{\textbf{Visual representation of the converged success rates in the MuJoCo setup.} A-B: The Median, IQM and Mean of the converged success rates (bootstrap replication 1000) for each Configuration in the sensorized and control groups, respectively. Confidence intervals are represented as shaded areas. Each configuration was run with 10 seeds; the exact values represented here can be found in Table \ref{tab:table2}.}
  \label{fig:fig6}
\end{figure}

Therefore, to better analyze these results, we should look at the converged success rates using bootstrapping, shown in Figure \ref{fig:fig6} and Table \ref{tab:table2}. The IQM of the converged success rate for the main group using Configuration 1 was only 0.8\% higher than the control, and its CI covered in its entirety the CI of the control group. The main group of Configuration 2, however, performed with a 7.5\% higher success rate than its control, with a 65\% overlap in CIs. The IQM success rate using Configuration 3 was the same for both main and control groups, with a 79\% overlap in CIs. With the same amount of overlap, the main group using Configuration 4 achieved a 0.8\% lower success rate than its control. The main group of Configuration 5 had a decidedly lower performance than the control group, with a success rate 12.5\% lower than the control, and only 42\% overlap between CIs. Finally, the control group, using Configuration 6, performed with a success rate 1.7\% higher than the main group, with 100\% of its CI overlapping.\\
When comparing the results of the main groups in the MuJoCo environment, we can see that Configuration 1 did not perform as similarly to other configurations as in the PyBullet environment. With a success rate of 84\% and a CI of 23\%, it underperformed relative to Configurations 2 and 3. Configuration 2, with fewer tactile sensors, achieved the highest success rate among all groups, including controls, with an 88\% success rate and a 17\% CI. Configuration 3, with the missing central sensor in this setup, performed with a similar success rate and CI range as Configuration 2, 87\% and 16\%, respectively. Configuration 4, which had the same number of sensors as Configuration 2, had a lower success rate with 84\%, and a similar CI, with 18\%, showing again how the different sensor layout affects learning. Configurations 5 and 6, which had a single central sensor, performed worse than other configurations, both in terms of success rate and CI. Configuration 5, with the small central sensor, performed worse with only a 73\% success rate, paired with a CI of 28\%. Compared to this, Configuration 6 did better, with a 78\% success rate and a CI of 30\%.

When looking at the results from both setups side-by-side (still analyzing IQMs), we can see that the main difference is in the average values for success rates and CIs. Although the MuJoCo setup resulted in higher success rates on average, it also generated larger CIs and overlaps between them. In both setups, Configurations 1 and 2 had higher success rates with sensory input, whereas Configurations 4 and 5 had higher success rates for the controls. Configuration 5 had the lowest performance, achieving the lowest success rate and high CI using sensors. Configuration 2, on the other hand, achieved the highest success rate in both setups, with a relatively small CI, compared to other configurations in the setup. The same configuration also achieved the highest lower and upper brackets of CIs in either setup. Another phenomenon that remained the same in both cases was the different performances of Configurations 2 and 4. These two had the same number of tactile sensors installed, but were positioned to produce different layouts, both of which included a central sensor. With their results, we can see how a different layout alone affects learning, as Configuration 4 had a 4-7\% worse success rate, depending on the setup, without a change in the size of CIs. 

\section{Discussion}
In this paper, we presented results on how different tactile sensor densities and layouts affect the performance of a machine learning agent when executing a grasp-and-lift task using a 3D hand model. With increasing interest in tactile sensors, new research is needed to investigate how this new modality may be integrated into existing robotic control schemes. These experiments aimed to support the use of tactile sensors in robotic hand control schemes by investigating their effects in simulated environments. We created two separate setups in two of the most commonly used simulators: PyBullet and MuJoCo. The different setups not only ensure that our results are robust, it also makes them relevant to a wider audience. The setup in PyBullet was made with the Modular Prosthetic Limb by Johns Hopkins \cite{johannes_chapter_2020} and a Proximal Policy Optimization \cite{schulman_proximal_2017} algorithm, which has been used increasingly in machine learning. This setup yielded results with fairly small variations, providing a clear insight into the positive effect of tactile information as well as showcasing how different sensor densities and layouts affect learning. The MuJoCo setup, on the other hand, used the Shadow Dexterous hand \cite{shadowrobot2013}, a widely used hand model, and a combination of Deep Deterministic Policy Gradient and Hindsight Experience Replay \cite{andrychowicz_hindsight_2018}, which yielded, on average, higher success rates, but also gave more variable results, as a whole. \\
All generated data were analyzed following the methods proposed by Agarwal et al. \cite{agarwal_deep_2022}, using the  RLiable Python library and Melnik et al. \cite{melnik_using_2021} to calculate converged success rates. Both setups used 6 different sensor layouts with varying densities, with control run type versions matching the number of sensors, but returning no tactile information. With the task being considered completed if the grabbed object is held securely above the ground, the agents learned how to pick up a cube with no visual feedback, only proprioceptive, or proprioceptive and tactile data. Our results have shown that, although the task is learnable even without tactile information, including it resulted in better performance using Configurations 1-3 and 6 in the PyBullet setup and Configurations 1-2 in the MuJoCo setup. In both setups, Configuration 2 performed the best out of all groups and configurations. This is an especially influential finding, as it performed better than Configuration 1, which had a higher density of sensors. This aligns with the findings of Melnik et al. \cite{melnik_using_2021}, namely that there is a certain density, after which performance no longer increases with the number of tactile sensors. This also suggests that a robotic hand does not need to be covered maximally with tactile pads to benefit from their addition; some space can be reserved for other modalities without sacrificing performance. We also showed that two configurations with matching sensor numbers performed differently, with Configuration 2 outperforming Configuration 4, where the only distinction was the layout of sensors. Although our results already show some effects that persist across different setups, they also highlight the benefits of further exploring this subject. With the variability of tactile sensors, virtually any layout and density may be simulated and tested, decreasing the amount of individual experimentation needed for robotic hand development. Utilizing additional learning algorithms, control schemes, and hand models would also shine light on which effects are robust across certain or even all setups. 

We identified several unexplored directions that may further expand on our experiments, but did not have the capacity to address. Our tuning of the learning algorithms extended only as much as it was needed to ensure that learning occurred, but no optimization was conducted. For example, each configuration was taught with the same neural network, which means that the size of the network may have been optimal for some configurations but sub-optimal for others. We hoped to visualize these differences through the control-type runs, as the results of these versions should only be affected by how much the neural network size is compatible with the size of the observation matrices. We also debated about the proper way to define the control type setups. Keeping the observation matrices the same by including an all-zero sensor output seemed to match the investigative nature of our work best, whereas deducting the sensor data altogether seemed the most realistic choice. For the sake of easy comparisons, we chose to include the all-zero sensor data; however, we acknowledge that the complete absence of it would be a viable alternative as well, especially when looking at a sim-to-real transfer of these experiments. Our future work involves introducing other objects into the setup to examine the effects of tactile sensors on learning to lift objects of different shapes, which may require inherently different grasp types. We are also working towards equipping the Shadow Hand model with a soft covering on the palmar side, as an additional, passive tool to aid in the formation and maintenance of secure grips on objects. Finally, we are also considering the implementation of a second version of control-type runs that would not include the zeroed sensor data to better cater to real-life implementation methods.

\section{Methods}
\subsection{PyBullet setup}
In the PyBullet \cite{pybulletcoumans2016} setup, we used the Modular Prosthetic Limb (MPL, \cite{johannes_chapter_2020}), which originally had 26 DoF when looking at the whole arm. From this model, we only used the hand, that was modified to allow for the continuous movement of each of the fingers along a given trajectory through PID control and to allow movement for the whole hand in all three axes of the 3D space. The fixed trajectories of the fingers were implemented to reduce the size of the action space, and facilitate learning. Each finger can be used to bend or straighten, but individual joint positions cannot be set. We also found it necessary to couple the actions of the thumb and little fingers, as this seemed to be the cornerstone that allowed for learning to begin. With that, possible actions of the hand model included: hand movements along the x, y and z axes, fingers 1 and 5 bending together, finger 2 bending, finger 3 bending and finger 4 bending, creating an action space of size 7. State observation included the object ID, Cartesian positions of both the robot and the object, as well as the states of the first joints of the five fingers. It also included the touch-sensor output (139/79/64/79/19/19 based on sensor configurations, as described below), which was set to zero in the control versions to ensure a consistent observation matrix size across the run types for each configuration.\\
The custom-made tactile sensors returned 1 if a contact point between the hand model and the environment (excluding the ground) fell within their sensing area, and 0 if it did not. On the phalanges, the sensor radius, the spaces between the rows and columns were 2 mm, 4.5 mm, 6.5 mm, respectively. (Please note that the default unit of the XML file format is in meters, and values should be adjusted accordingly.) Sensors on the palm had the same radius and were positioned at the base of the index, middle, ring, and little fingers. Figure \ref{fig:fig2} visualizes the 6 Configurations in which the tactile sensors were arranged. For Configuration 6, the sensor radius was set to 3.8 mm. Please note that the pattern shown on the index finger were repeated on all fingers, resulting in a total of 139, 79, 64, 79, 19 and 19 tactile sensors in Configurations 1-6, respectively.

\subsubsection{Reinforcement learning algorithm}
Learning in the PyBullet setup is carried out by a Proximal Policy Optimization \cite{schulman_proximal_2017, yu2023ppo} (PPO) algorithm. The main parameters of the implemented version of PPO were as follows:
\begin{itemize}
    \item Actor network: 3$\times$256, lr = $3\times 10^{-4}$, ReLU, Tanh output
    \item Critic network: 3$\times$256, lr = $1\times 10^{-3}$, ReLU, Identity output
    \item Adam optimizer \cite{kingma_adam_2017}
    \item Action distribution: Gaussian, diagonal covariance ($\sigma^2 = 0.05$)
    \item Discount factor: $\gamma = 0.99$
    \item Advantage estimator: GAE ($\lambda = 0.97$)
    \item Clip range: 0.2
    \item Minibatch size: 6
    \item Max. timesteps per episode: 100
    \item Timesteps per epoch: 1000
    \item Total timesteps: 500000
    \item Updates per epoch: 80
    \item Entropy coefficient: 0.0
    \item Gradient clipping: 0.5
\end{itemize}
The learned policy outputs an action vector of size 7 to control the hand and its fingers, which contains one pair of coupled digits (thumb and little fingers). The values range from -1 to 1, which are implemented directly, in relation to the current position and within the limits of the hand model. Episodes begin with partially closed fingers, and the object is placed directly under the hand model. The task requires the hand to grab and lift the object, keeping it securely in the air for 5000 unrecorded simulation steps. The reward function is one of the most crucial parts of a PPO algorithm. In our case, the reward function was built to return different values based on task progression. While the object has not been moved from the ground, the agent receives a reward based on the closeness of the hand and fingers to the object center. 
\[
R = \exp\Big(-\|\mathbf{hand} - \mathbf{obj}\|^2\Big) 
    + \alpha \sum_{i=1}^{5} \exp\Big(-\|\mathbf{f}_i - \mathbf{obj}\|^2\Big)
\]

where \(\alpha\)=0.2, \textbf{hand} and \textbf{obj} are the Cartesian coordinates of the hand and the object, respectively, and \textbf{f} denotes the Cartesian coordinates of the distal phalanges. Alpha is used to scale the rewards from the fingers so that their maximal value cannot exceed 1. If there was a failed attempt at lifting the object off the ground, the agent will receive an additional reward of 5. If the object was lifted into the air successfully, the hand is moved to a height of 1 meter and kept there for 5000 unrecorded simulation steps. After this, if the object is still in the air, the agent is given a reward of 1000, and the episode ends. If the object is dropped, it will still receive the extra reward of 5, and the episode ends. If the episode reaches the maximal timesteps per episode, the episode ends, and the agent receives a reward of -100. We found that this reward scheme encourages object manipulation without enabling the option to get high rewards without completing the task.

\subsection{MuJoCo setup}
In the MuJoCo \cite{todorov_mujoco_2012} setup, we used the Shadow Dexterous Hand (Shadow Hand) \cite{shadowrobot2013} from the codebase of Melnik et al. \cite{melnik_using_2021}. This model originally had 24 DoF, which we extended to allow movement in 3D space. Possible actions included joint values for the five fingers (20) and hand movement along the x,y, and z axes to create an action space of size 23. State observation included the angle values and velocities of the robot’s joints, as well as the Cartesian position and rotation of the object. It also included the touch-sensor output (139/79/64/79/19/19 based on sensor configurations, as described below), which was set to zero in the control versions to ensure a consistent observation matrix size across runs for each configuration.\\
The touch sensors in this setup are built-in MuJoCo touch sensors that return Boolean values for each query. Due to variations in phalange size, the sensor radius and inter-column spacing were fixed at 1.5 mm and 5 mm in this setup, whereas the inter-row spacing was set to 6 mm, 3 mm, and 7.5 mm for the proximal, middle, and distal phalanges, respectively. Sensors on the palm had the same radius and were positioned at the base of the index, middle, ring, and little fingers. Figure \ref{fig:fig2} visualizes the 6 Configurations in which the tactile sensors were arranged. For Configuration 6, the radius of the sensor was set to 3.5 mm. Please note that the pattern shown on the index finger were repeated on all fingers, resulting in a total of 139, 79, 64, 79, 19 and 19 tactile sensors in Configurations 1-6, respectively.

\subsubsection{Reinforcement learning algorithm}
Learning in the MuJoCo setup is carried out by a combination of Deep Deterministic Policy Gradient \cite{lillicrap_continuous_2019} (DDPG) and Hindsight Experience Replay \cite{andrychowicz_hindsight_2018} (HER) algorithms as described by Melnik et al. \cite{melnik_using_2021} (2021) and Plappert et al. \cite{plappert_multi-goal_2018} (2018). The hyperparameters were as follows \cite{melnik_using_2021}:
\begin{itemize}
    \item Actor and critic networks: 3 layers with 256 units each and ReLU non-linearities
    \item Adam optimizer \cite{kingma_adam_2017} with $10^{-3}$ for training both actor and critic
    \item Buffer size: $10^{6}$ transitions
    \item Polyak-averaging coefficient: 0.95
    \item Action L2 norm coefficient: 1.0
    \item Observation clipping: $[-200, 200]$
    \item Batch size: 256
    \item Rollouts per MPI worker: 2
    \item Number of MPI workers: 19
    \item Total timesteps: 5000000
    \item Cycles per epoch: 50
    \item Batches per cycle: 40
    \item Test rollouts per epoch: 10
    \item Probability of random actions: 0.3
    \item Scale of additive Gaussian noise: 0.2
    \item Probability of HER experience replay: 0.8
    \item Normalized clipping: $[-5, 5]$
\end{itemize}

The policy output includes 23 continuous values in the range of -1 and 1 to control the 23 actuated joints of the hand model. These normalized values are converted to joint positions using linear mapping  (actuation center + action × actuation range/2) \cite{melnik_using_2021}. Episodes begin with partially closed fingers, and the object is placed directly under the hand model. The goal position is always located above the hand, so that no part of the object or the hand may touch the ground when reached. The task requires the hand to grab and move the object to the target location and keep it there until the end of the episode. Rewards are given at each timestep according to the distance between the object's current and desired locations, with a scaling of 10. Rotation of the object is not considered in neither the reward shaping nor goal achievement. A reward of 0 is given if the object is delivered to the target location, with some tolerance \cite{melnik_using_2021}.  A reward of -1 is given if the target has not been reached by the end of the episode. Both the starting and goal locations (and orientations) are set to be the same across all runs. Simulations were run with 2 parallel environments per MPI worker.

\section*{Acknowledgements}
This work was supported by the Ministry of Culture and Innovation of Hungary from the National Research, Development and Innovation Fund, financed under the TKP2021-NKTA funding scheme (project no. TKP2021-NKTA-66) and under the KDP-2023 funding scheme (project no. 2023-2.1.2-KDP-2023-00011 C2299141). We would also like to thank the contributions of Domonkos Körmendy and Jedlik Innovation LLC to the realization of this paper.

\section*{Author contributions}
M.K. and E.B. formed the underlying idea and design of the research. With the supervision of M.K. E.B. developed the code, ran the simulations and analyzed the results. E.B. wrote the manuscript, and M.K. reviewed and consulted on its content.

\section*{Competing interests}
All authors declare no financial or non-financial competing interests.

\section*{Data availability}
The data generated and analysed during the current study will be made available in a Zonedo repository.

\section*{Code availability}
The underlying code for this study is not publicly available but may be made available to qualified researchers on reasonable request from the corresponding author.
\vspace{0.5cm}
\small
\printbibliography

@article{shi_computer_2020,
	title = {Computer Vision-Based Grasp Pattern Recognition With Application to Myoelectric Control of Dexterous Hand Prosthesis},
	volume = {28},
	doi = {10.1109/TNSRE.2020.3007625},
	pages = {2090--2099},
	number = {9},
	journaltitle = {{IEEE} Transactions on Neural Systems and Rehabilitation Engineering},
	author = {Shi, Chunyuan and Yang, Dapeng and Zhao, Jingdong and Liu, Hong},
	date = {2020-09},
}

@article{zhao_embedding_2025,
	title = {Embedding high-resolution touch across robotic hands enables adaptive human-like grasping},
	volume = {7},
	doi = {10.1038/s42256-025-01053-3},
	pages = {889--900},
	number = {6},
	journaltitle = {Nat Mach Intell},
	author = {Zhao, Zihang and Li, Wanlin and Li, Yuyang and Liu, Tengyu and Li, Boren and Wang, Meng and Du, Kai and Liu, Hangxin and Zhu, Yixin and Wang, Qining and Althoefer, Kaspar and Zhu, Song-Chun},
	date = {2025-06},
}

@inproceedings{xu_towards_2022,
	title = {Towards Learning to Play Piano with Dexterous Hands and Touch},
	doi = {10.1109/IROS47612.2022.9981221},
	pages = {10410--10416},
	booktitle = {2022 {IEEE}/{RSJ} International Conference on Intelligent Robots and Systems ({IROS})},
	author = {Xu, Huazhe and Luo, Yuping and Wang, Shaoxiong and Darrell, Trevor and Calandra, Roberto},
	date = {2022-10},
}

@article{andrychowicz_learning_2020,
	title = {Learning dexterous in-hand manipulation},
	volume = {39},
	doi = {10.1177/0278364919887447},
	pages = {3--20},
	number = {1},
	journaltitle = {The International Journal of Robotics Research},
	author = {Andrychowicz, {OpenAI}: Marcin and Baker, Bowen and Chociej, Maciek and Józefowicz, Rafal and {McGrew}, Bob and Pachocki, Jakub and Petron, Arthur and Plappert, Matthias and Powell, Glenn and Ray, Alex and Schneider, Jonas and Sidor, Szymon and Tobin, Josh and Welinder, Peter and Weng, Lilian and Zaremba, Wojciech},
	date = {2020-01-01},
}

@article{melnik_using_2021,
	title = {Using Tactile Sensing to Improve the Sample Efficiency and Performance of Deep Deterministic Policy Gradients for Simulated In-Hand Manipulation Tasks},
	volume = {8},
	doi = {10.3389/frobt.2021.538773},
	journaltitle = {Front. Robot. {AI}},
	author = {Melnik, Andrew and Lach, Luca and Plappert, Matthias and Korthals, Timo and Haschke, Robert and Ritter, Helge},
	date = {2021-06-29},
}

@article{cotton_novel_2007,
	title = {A Novel Thick-Film Piezoelectric Slip Sensor for a Prosthetic Hand},
	volume = {7},
	doi = {10.1109/JSEN.2007.894912},
	pages = {752--761},
	number = {5},
	journaltitle = {{IEEE} Sensors Journal},
	author = {Cotton, Darryl P. J. and Chappell, Paul H. and Cranny, Andy and White, Neil M. and Beeby, Steve P.},
	date = {2007-05},
}

@article{cipriani_smarthand_2011,
	title = {The {SmartHand} transradial prosthesis},
	volume = {8},
	doi = {10.1186/1743-0003-8-29},
	pages = {29},
	number = {1},
	journaltitle = {Journal of {NeuroEngineering} and Rehabilitation},
	author = {Cipriani, Christian and Controzzi, Marco and Carrozza, Maria Chiara},
	date = {2011-05-22},
}

@article{yuan_gelsight_2017,
	title = {{GelSight}: High-Resolution Robot Tactile Sensors for Estimating Geometry and Force},
	volume = {17},
	doi = {10.3390/s17122762},
	pages = {2762},
	number = {12},
	journaltitle = {Sensors},
	author = {Yuan, Wenzhen and Dong, Siyuan and Adelson, Edward H.},
	date = {2017-12},
}

@article{james_slip_2018,
	title = {Slip Detection With a Biomimetic Tactile Sensor},
	volume = {3},
	doi = {10.1109/LRA.2018.2852797},
	pages = {3340--3346},
	number = {4},
	journaltitle = {{IEEE} Robotics and Automation Letters},
	author = {James, Jasper Wollaston and Pestell, Nicholas and Lepora, Nathan F.},
	date = {2018-10},
}

@incollection{Maggiali08,
    author = {Giorgio Cannata and Marco Maggiali},
    title = {Design of a Tactile Sensor for Robot Hands},
    booktitle = {Sensors - Focus on Tactile Force and Stress Sensors},
    publisher = {IntechOpen},
    year = {2008},
    editor = {Jose Gerardo Rocha and Senentxu Lanceros-Mendez},
    chapter = {14},
    doi = {10.5772/6626},
}

@article{zhang_design_2018,
	title = {Design and Functional Evaluation of a Dexterous Myoelectric Hand Prosthesis With Biomimetic Tactile Sensor},
	volume = {26},
	doi = {10.1109/TNSRE.2018.2844807},
	pages = {1391--1399},
	number = {7},
	journaltitle = {{IEEE} Trans Neural Syst Rehabil Eng},
	author = {Zhang, Ting and Jiang, Li and Liu, Hong},
	date = {2018-07},
	pmid = {29985148},
}

@article{valarezo_anazco_natural_2021,
	title = {Natural object manipulation using anthropomorphic robotic hand through deep reinforcement learning and deep grasping probability network},
	volume = {51},
	doi = {10.1007/s10489-020-01870-6},
	pages = {1041--1055},
	number = {2},
	journaltitle = {Appl Intell},
	author = {Valarezo Añazco, Edwin and Rivera Lopez, Patricio and Park, Nahyeon and Oh, Jiheon and Ryu, Gahyeon and Al-antari, Mugahed A. and Kim, Tae-Seong},
	date = {2021-02-01},
	langid = {english},
}

@article{kampmann_integration_2014,
	title = {Integration of Fiber-Optic Sensor Arrays into a Multi-Modal Tactile Sensor Processing System for Robotic End-Effectors},
	volume = {14},
	doi = {10.3390/s140406854},
	pages = {6854--6876},
	number = {4},
	journaltitle = {Sensors},
	author = {Kampmann, Peter and Kirchner, Frank},
	date = {2014-04},
}

@inproceedings{castellanos-ramos_tactile_2009,
	title = {Tactile sensors based on conductive polymers},
	volume = {7362},
	doi = {10.1117/12.821627},
	pages = {140--148},
	booktitle = {Smart Sensors, Actuators, and {MEMS} {IV}},
	author = {Castellanos-Ramos, Julian and Navas-Gonzalez, Rafael and Macicior, Haritz and Ochoteco, Estibalitz and Vidal-Verdú, Fernando},
	date = {2009-05-18},
}

@article{abbass_embedded_2021,
	title = {Embedded Electrotactile Feedback System for Hand Prostheses Using Matrix Electrode and Electronic Skin},
	volume = {15},
	doi = {10.1109/TBCAS.2021.3107723},
	pages = {912--925},
	number = {5},
	journaltitle = {{IEEE} Transactions on Biomedical Circuits and Systems},
	author = {Abbass, Yahya and Saleh, Moustafa and Dosen, Strahinja and Valle, Maurizio},
	date = {2021-10},
}

@inproceedings{navarai_capacitive-piezoelectric_2018,
	title = {Capacitive-Piezoelectric Tandem Architecture for Biomimetic Tactile Sensing in Prosthetic Hand},
	doi = {10.1109/ICSENS.2018.8589827},
	pages = {1--4},
	booktitle = {{IEEE} Sensors Journal},
	author = {Navarai, William Taube and Ozioko, Oliver and Dahiya, Ravinder},
	date = {2018-10},
}

@inproceedings{osborn_tactile_2014,
	title = {Tactile feedback in upper limb prosthetic devices using flexible textile force sensors},
	doi = {10.1109/BIOROB.2014.6913762},
	pages = {114--119},
	booktitle = {5th {IEEE} {RAS}/{EMBS} International Conference on Biomedical Robotics and Biomechatronics},
	author = {Osborn, Luke and Lee, Wang Wei and Kaliki, Rahul and Thakor, Nitish},
	date = {2014-08},
}

@inproceedings{navaraj_prosthetic_2019,
	title = {Prosthetic Hand with Biomimetic Tactile Sensing and Force Feedback},
	doi = {10.1109/ISCAS.2019.8702732},
	pages = {1--4},
	booktitle = {2019 {IEEE} International Symposium on Circuits and Systems ({ISCAS})},
	author = {Navaraj, William Taube and Nassar, Habib and Dahiya, Ravinder},
	date = {2019-05},
}

@article{sundaralingam_relaxed-rigidity_2017,
	title = {Relaxed-rigidity constraints: In-grasp manipulation using purely kinematic trajectory optimization},
	volume = {13},
	doi = {10.15607/rss.2017.xiii.015},
	journaltitle = {Robotics: Science and Systems},
	author = {Sundaralingam, Balakumar and Hermans, Tucker},
	date = {2017},
}

@article{schmitz_methods_2011,
	title = {Methods and Technologies for the Implementation of Large-Scale Robot Tactile Sensors},
	volume = {27},
	doi = {10.1109/TRO.2011.2132930},
	pages = {389--400},
	number = {3},
	journaltitle = {{IEEE} Transactions on Robotics},
	author = {Schmitz, Alexander and Maiolino, Perla and Maggiali, Marco and Natale, Lorenzo and Cannata, Giorgio and Metta, Giorgio},
	date = {2011-06},
}

@article{fukui_high-speed_2011,
	title = {High-Speed Tactile Sensing for Array-Type Tactile Sensor and Object Manipulation Based on Tactile Information},
	volume = {2011},
	doi = {10.1155/2011/691769},
	pages = {691769},
	number = {1},
	journaltitle = {Journal of Robotics},
	author = {Fukui, Wataru and Kobayashi, Futoshi and Kojima, Fumio and Nakamoto, Hiroyuki and Imamura, Nobuaki and Maeda, Tadashi and Shirasawa, Hidenori},
	date = {2011},
}

@inproceedings{mouri_anthropomorphic_2002,
    author = {Mouri, Tetsuya and Kawasaki, Haruhisa and Yoshikawa, K. and Takai, J. and Ito, Soken},
    year = {2002},
    month = {01},
    pages = {1288-1293},
    title = {Anthropomorphic robot hand: Gifu hand III},
    booktitle = {Proc. of Int. Conf. ICCAS2002}
}

@misc{plappert_multi-goal_2018,
	title = {Multi-Goal Reinforcement Learning: Challenging Robotics Environments and Request for Research},
	doi = {10.48550/arXiv.1802.09464},
	number = {{arXiv}:1802.09464},
	author = {Plappert, Matthias and Andrychowicz, Marcin and Ray, Alex and {McGrew}, Bob and Baker, Bowen and Powell, Glenn and Schneider, Jonas and Tobin, Josh and Chociej, Maciek and Welinder, Peter and Kumar, Vikash and Zaremba, Wojciech},
	date = {2018-03-10},
	eprinttype = {arxiv},
	eprint = {1802.09464 [cs]},
}

@misc{yu2023ppo,
  author       = {Eric Yu},
  title        = {PPO-for-Beginners: A simple and well-styled PPO implementation},
  howpublished = {\url{https://github.com/ericyangyu/PPO-for-Beginners}},
  year         = {2023},
  note         = {GitHub repository}
}

@inproceedings{pybulletcoumans2016,
  title={PyBullet, a python module for physics simulation for games, robotics and machine learning},
  author={Coumans, Erwin and Bai, Yunfei},
  booktitle={http://pybullet.org},
  year={2016}
}

@misc{kingma_adam_2017,
	title = {Adam: A Method for Stochastic Optimization},
	doi = {10.48550/arXiv.1412.6980},
	number = {{arXiv}:1412.6980},
	publisher = {{arXiv}},
	author = {Kingma, Diederik P. and Ba, Jimmy},
	date = {2017-01-30},
	eprinttype = {arxiv},
	eprint = {1412.6980 [cs]},
}

@inproceedings{todorov_mujoco_2012,
	title = {{MuJoCo}: A physics engine for model-based control},
	doi = {10.1109/IROS.2012.6386109},
	pages = {5026--5033},
	booktitle = {2012 {IEEE}/{RSJ} International Conference on Intelligent Robots and Systems},
	author = {Todorov, Emanuel and Erez, Tom and Tassa, Yuval},
	date = {2012-10},
}

@misc{andrychowicz_hindsight_2018,
	title = {Hindsight Experience Replay},
	doi = {10.48550/arXiv.1707.01495},
	number = {{arXiv}:1707.01495},
	publisher = {{arXiv}},
	author = {Andrychowicz, Marcin and Wolski, Filip and Ray, Alex and Schneider, Jonas and Fong, Rachel and Welinder, Peter and {McGrew}, Bob and Tobin, Josh and Abbeel, Pieter and Zaremba, Wojciech},
	date = {2018-02-23},
	eprinttype = {arxiv},
	eprint = {1707.01495 [cs]},
}

@misc{schulman_proximal_2017,
	title = {Proximal Policy Optimization Algorithms},
	doi = {10.48550/arXiv.1707.06347},
	number = {{arXiv}:1707.06347},
	publisher = {{arXiv}},
	author = {Schulman, John and Wolski, Filip and Dhariwal, Prafulla and Radford, Alec and Klimov, Oleg},
	date = {2017-08-28},
	eprinttype = {arxiv},
	eprint = {1707.06347 [cs]},
}

@misc{lillicrap_continuous_2019,
	title = {Continuous control with deep reinforcement learning},
	doi = {10.48550/arXiv.1509.02971},
	number = {{arXiv}:1509.02971},
	publisher = {{arXiv}},
	author = {Lillicrap, Timothy P. and Hunt, Jonathan J. and Pritzel, Alexander and Heess, Nicolas and Erez, Tom and Tassa, Yuval and Silver, David and Wierstra, Daan},
	date = {2019-07-05},
	eprinttype = {arxiv},
	eprint = {1509.02971 [cs]},
}

@misc{shadowrobot2013,
  author       = {Shadow Robot Company},
  title        = {Shadow Dexterous Hand Technical Specification},
  howpublished = {\url{https://www.shadowrobot.com/products/dexterous-hand}},
  year         = {2013}
}

@incollection{johannes_chapter_2020,
	title = {Chapter 21 - The Modular Prosthetic Limb},
	pages = {393--444},
	booktitle = {Wearable Robotics},
	publisher = {Academic Press},
	author = {Johannes, Matthew S. and Faulring, Eric L. and Katyal, Kapil D. and Para, Matthew P. and Helder, John B. and Makhlin, Alexander and Moyer, Tom and Wahl, Daniel and Solberg, James and Clark, Steve and Armiger, Robert S. and Lontz, Travis and Geberth, Kathryn and Moran, Courtney W. and Wester, Brock A. and Van Doren, Thomas and Santos-Munne, Julio J.},
	editor = {Rosen, Jacob and Ferguson, Peter Walker},
	date = {2020-01-01},
	doi = {10.1016/B978-0-12-814659-0.00021-7},
}

@inproceedings{yu_realtime_2018,
	title = {Realtime State Estimation with Tactile and Visual Sensing. Application to Planar Manipulation},
	doi = {10.1109/ICRA.2018.8463183},
	pages = {7778--7785},
	booktitle = {2018 {IEEE} International Conference on Robotics and Automation ({ICRA})},
	author = {Yu, Kuan–Ting and Rodriguez, Alberto},
	date = {2018-05},
}

@inproceedings{motamedi_haptic_2016,
	title = {Haptic feedback for improved robotic arm control during simple grasp, slippage, and contact detection tasks},
	doi = {10.1109/ICRA.2016.7487694},
	pages = {4894--4900},
	booktitle = {2016 {IEEE} International Conference on Robotics and Automation ({ICRA})},
	author = {Motamedi, M. Reza and Chossat, Jean-Baptiste and Roberge, Jean-Philippe and Duchaine, Vincent},
	date = {2016-05},
}

@misc{agarwal_deep_2022,
	title = {Deep Reinforcement Learning at the Edge of the Statistical Precipice},
	doi = {10.48550/arXiv.2108.13264},
	number = {{arXiv}:2108.13264},
	publisher = {{arXiv}},
	author = {Agarwal, Rishabh and Schwarzer, Max and Castro, Pablo Samuel and Courville, Aaron and Bellemare, Marc G.},
	date = {2022-01-05},
	eprinttype = {arxiv},
	eprint = {2108.13264 [cs]},
}
\section{Figure legends}

Figure 1. Schematic overview of the methodology used to evaluate tactile sensor configurations. The same configurations have been implemented on two 3D hand models using different physics simulators. The effects of the different sensor layouts were evaluated using two reinforcement learning algorithms to ensure the robustness of the results and compared against configurations, setups and control groups.

Figure 2. Visual representation of the sensor configurations implemented in both setups. A-F: Sensor configurations corresponding to Configurations 1-6, respectively. The number and layout of the sensors shown on the index finger are repeated across the other digits.

Figure 3. Sample-efficiency curves showing success rates as a function of epochs in the PyBullet setup. A-F: Sample-efficiency curves corresponding to Configurations 1-6 (shown on diagrams in the right hand corner of each panel), respectively. Thick lines show the IQM scores (bootstrap replication 50000), with the confidence intervals represented as shaded areas. Yellow is used to show results from the sensorized group, while blue is used for the controls. Training occurred for 500000 timesteps, roughly generating 500 epochs. Both main and control groups represent the result of 10 seeds.

Figure 4. Visual representation of the converged success rates in the PyBullet setup. A-B: The Median, IQM and Mean of the converged success rates (bootstrap replication 1000) for each Configuration in the sensorized and control groups, respectively. Confidence intervals are represented as shaded areas. Each configuration was run with 10 seeds; the exact values represented here can be found in Table 1.

Figure 5. Sample-efficiency curves showing success rates as a function of epochs in the MuJoCo setup.  A-F: Sample-efficiency curves corresponding to Configurations 1-6 (shown on diagrams in the right hand corner of each panel), respectively. Thick lines show the IQM scores (bootstrap replication 50000), with the confidence intervals represented as shaded areas. Yellow is used to show results from the sensorized group, while blue is used for the controls. Training occurred for 5000000 timesteps, generating 500 epochs. Both main and control groups represent the result of 10 seeds.

Figure 6. Visual representation of the converged success rates in the MuJoCo setup. A-B: The Median, IQM and Mean of the converged success rates (bootstrap replication 1000) for each Configuration in the sensorized and control groups, respectively. Confidence intervals are represented as shaded areas. Each configuration was run with 10 seeds; the exact values represented here can be found in Table 2.

\begin{table*}[h]
    \centering
    \begin{tabular}{cccccccc}
        Config. & Type & Median & IQM & Mean & Median CI, 95\%  & IQM  CI, 95\%  & Mean  CI, 95\% \\
        \rowcolor{Gray}
        1 & main & 0.65 & 0.65	& 0.65 & [0.62, 0.67] & [0.61, 0.68] & [0.62, 0.67]	\\
        1 & control & 0.60 & 0.62	& 0.60 & [0.52, 0.65] & [0.54, 0.66] & [0.52, 0.65]\\  
        \rowcolor{Gray}
        2 & main & 0.66 & 0.66	& 0.66 & [0.62, 0.68] & [0.63, 0.69] & [0.62, 0.68]	\\
        2 & control & 0.61 & 0.62	& 0.61 & [0.57, 0.65] & [0.57, 0.67] & [0.57, 0.65]	\\   
        \rowcolor{Gray}
        3 & main & 0.58 & 0.64	& 0.58 & [0.45, 0.68] & [0.50, 0.68] & [0.45, 0.68]	\\
        3 & control & 0.62 & 0.63	& 0.62 & [0.60, 0.65] & [0.60, 0.65] & [0.60, 0.65]	\\ 
        \rowcolor{Gray}
        4 & main & 0.59 & 0.59	& 0.59 & [0.56, 0.62] & [0.55, 0.63] & [0.56, 0.62]	\\
        4 & control & 0.63 & 0.64	& 0.63 & [0.60, 0.65] & [0.59, 0.65] & [0.60, 0.65]\\  
        \rowcolor{Gray}
        5 & main & 0.58 & 0.57	& 0.58 & [0.54, 0.62] & [0.52, 0.64] & [0.54, 0.62]	\\
        5 & control & 0.58 & 0.61	& 0.58 & [0.50, 0.63] & [0.53, 0.64] & [0.50, 0.63]	\\ 
        \rowcolor{Gray}
        6 & main & 0.64 & 0.64	& 0.64 & [0.61, 0.67] & [0.61, 0.68] & [0.61, 0.67]	\\
        6 & control & 0.56 & 0.60	& 0.56 & [0.44, 0.64] & [0.49, 0.64] & [0.44, 0.64]	\\        
    \end{tabular}
    \caption{\textbf{Converged success rates in the PyBullet setup.} Median, Interquartile Mean (IQM) and Mean values were derived from converged success rates considering the entire training period. Each configuration was run with 10 seeds.}
    \label{tab:table1}
\end{table*}

\begin{table*}[h]
    \centering
    \begin{tabular}{cccccccc}
        Config. & Type & Median & IQM & Mean & Median CI, 95\% & IQM CI, 95\% & Mean CI, 95\% \\
        \rowcolor{Gray}
        1 & main & 0.82 & 0.84 & 0.82 & [0.72, 0.90] & [0.70,  0.93] & [0.72, 0.90] \\
        1 & control & 0.85 & 0.83 & 0.85 & [0.80, 0.90] & [0.78, 0.90] & [0.80, 0.90] \\
        \rowcolor{Gray}
        2 & main & 0.88 & 0.88 & 0.88 & [0.82, 0.94] & [0.79, 0.96] & [0.82, 0.94] \\
        2 & control & 0.81 & 0.80 & 0.81 & [0.75, 0.86] & [0.74, 0.88] & [0.75, 0.86] \\
        \rowcolor{Gray}
        3 & main & 0.87 & 0.87 & 0.87 & [0.82, 0.93] & [0.79, 0.95] & [0.82, 0.93] \\
        3 & control & 0.83 & 0.87 & 0.83 & [0.70, 0.90] & [0.76, 0.92] & [0.70, 0.90] \\
        \rowcolor{Gray}
        4 & main & 0.80 & 0.84 & 0.80 & [0.70, 0.88] & [0.72, 0.90] & [0.70, 0.88] \\
        4 & control & 0.81 & 0.85 & 0.81 & [0.70, 0.90] & [0.70, 0.93] & [0.70, 0.90] \\
        \rowcolor{Gray}
        5 & main & 0.70 & 0.73 & 0.70 & [0.56, 0.81] & [0.56, 0.83] & [0.56, 0.81] \\
        5 & control & 0.84 & 0.86 & 0.84 & [0.76, 0.92] & [0.77, 0.93] & [0.76, 0.92] \\
        \rowcolor{Gray}
        6 & main & 0.73 & 0.78 & 0.73 & [0.55, 0.86] & [0.59, 0.89] & [0.55, 0.86] \\
        6 & control & 0.80 & 0.79 & 0.80 & [0.75, 0.85] & [0.73, 0.85] & [0.75, 0.85] \\       
    \end{tabular}
    \caption{\textbf{Converged success rates in the MuJoCo setup.} Median, Interquartile Mean (IQM) and Mean values were derived from converged success rates considering the entire training period. Each configuration was run with 10 seeds.}
    \label{tab:table2}
\end{table*}

\end{document}